\documentclass[10pt,twocolumn,letterpaper]{article}

\usepackage{iccv}
\usepackage{times}
\usepackage{epsfig}
\usepackage{graphicx}
\usepackage{amsmath}
\usepackage{amssymb}
\usepackage{pifont}
\usepackage{epstopdf}
\usepackage{booktabs}
\usepackage{multirow}
\usepackage{array, makecell, caption}

\usepackage[pagebackref=true,breaklinks=true,letterpaper=true,colorlinks,bookmarks=false]{hyperref}

\iccvfinalcopy 

\def\iccvPaperID{10002} 

\ificcvfinal\fi

\begin{document}


\title{Language-aware Multiple Datasets Detection Pretraining for DETRs}

\author{
Jing Hao\footnotemark[1]\quad Song Chen\thanks{Equal contribution.} \quad Xiaodi Wang \quad Shumin Han\\
Baidu VIS\\
{\tt\small \{haojing08, chensong03, wangxiaodi03, hanshumin\}@baidu.com}
}

\maketitle
\ificcvfinal\thispagestyle{empty}\fi

\begin{abstract}
Pretraining on large-scale datasets can boost the performance of object detectors while the annotated datasets for object detection are hard to scale up due to the high labor cost.
What we possess are numerous isolated filed-specific datasets, thus, it is appealing to jointly pretrain models across aggregation of datasets to enhance data volume and diversity. 
In this paper, we propose a strong framework for utilizing \textbf{M}ultiple datasets to pretrain D\textbf{ETR}-like detectors, termed \textbf{METR}, without the need for manual label spaces integration.
%
%
It converts the typical multi-classification in object detection into binary classification by introducing a pre-trained language model.
Specifically, we design a category extraction module for extracting potential categories involved in an image and assign these categories into different queries by language embeddings. Each query is only responsible for predicting a class-specific object. Besides, to adapt our novel detection paradigm, we propose a group bipartite matching strategy that limits the ground truths to match queries assigned to the same category.
%
Extensive experiments demonstrate that METR achieves extraordinary results on either multi-task joint training or the pretrain \& finetune paradigm. Notably, our pre-trained models have high flexible transferability and increase the performance upon various DETR-like detectors on COCO val2017 benchmark. Codes will be available after this paper is published.
\end{abstract}

\begin{figure}
\centerline{\includegraphics[width=0.48\textwidth]{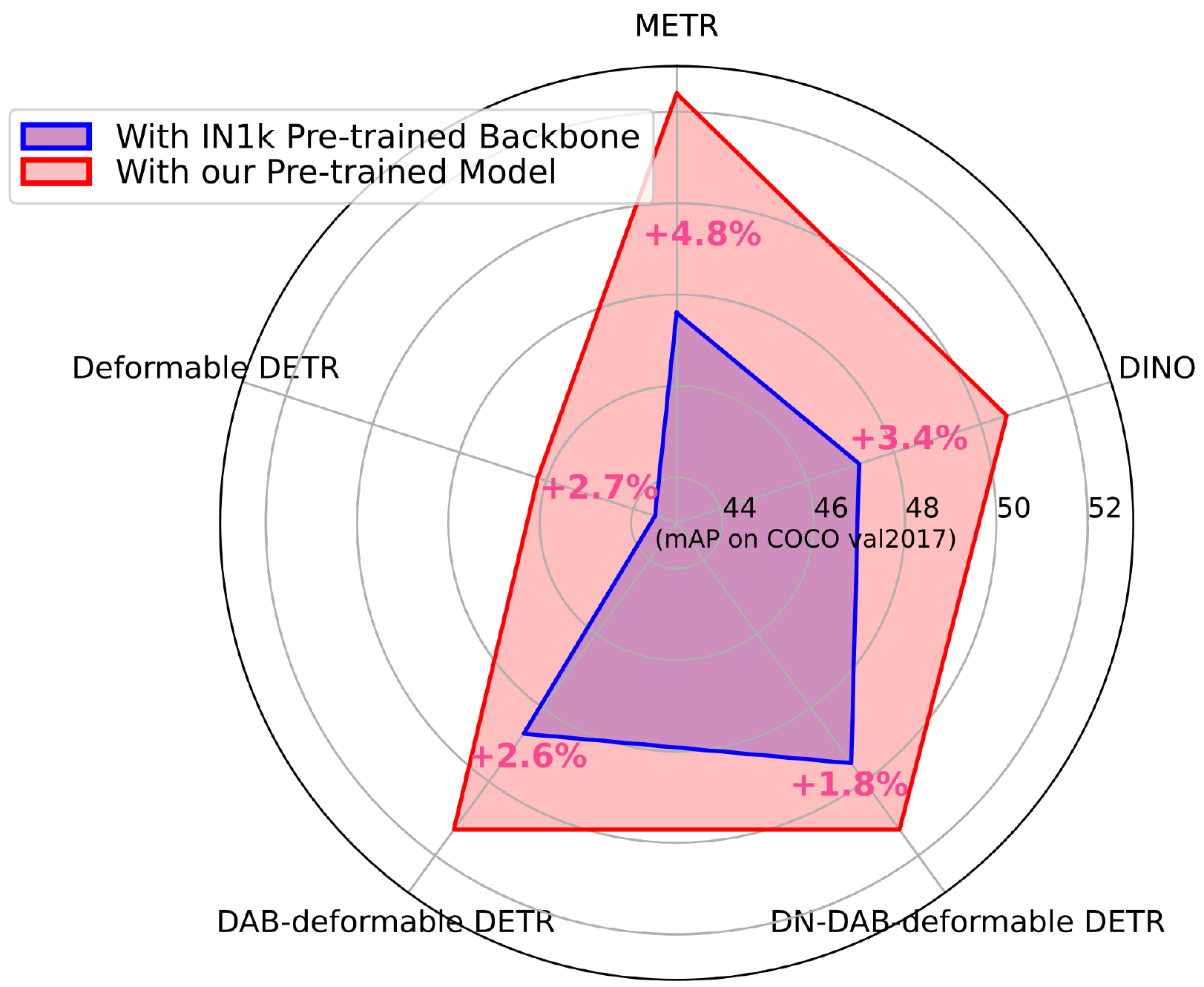}}
\caption{mAP on COCO val2017 compared with DETR-like detectors. Our pre-trained model significantly boosts the performance of DETRs with higher mean average precision. Note that METR and DINO \cite{zhang2022dino} are trained with 12 epochs, and the others are trained with 50 epochs.}
\label{fig:short}
\vspace{-6pt}
\end{figure}

\section{Introduction}

Object detection (OD), a fundamental computer vision task aiming to localize objects with tight bounding boxes in images, has been dominated by DETR-family detectors recently\cite{carion2020end,dai2021dynamic,meng2021conditional,zhudeformable}. These detectors eliminate the need for hand-designed components and achieve or even surpass the performance of optimized classical detectors like \cite{cai2018cascade,tian2019fcos,ge2021yolox,wang2022yolov7,xu2022pp}. Despite achieving promising performance, they inevitably suffer from slow convergence on training \cite{carion2020end}. Recent many methods \cite{dai2021dynamic,meng2021conditional,zhudeformable} have been proposed to accelerate the training convergence, especially in algorithm designing. For example, DN-DETR \cite{li2022dn} feeds noised ground truth bounding boxes into the decoder and trains the model to reconstruct the original boxes.

At the same time, the pre-training \& finetune paradigm has broadly applied to various visual recognition tasks because loading a pre-trained model usually can boost training convergence and performance \cite{arnab2021vivit,chen2019med3d,tajbakhsh2016convolutional,he2022masked}.
The mainstream pretraining algorithms can be split into two tracks according to the constructions of pre-trained models: 1) Pretraining for a visual backbone. Self-supervised learning methods \cite{chen2022context,he2022masked,he2020momentum} aim to pretrain a visual backbone with rich semantic representation, while the other parts of detectors designed for downstream tasks are ignored and usually initialized randomly. Take DETR as an example, the CNN backbone (ResNet-50 \cite{he2016deep} with 23.2M parameters) has been pretrained to extract a good visual representation, but the transformer modules with 18.0M parameters have not been pretrained.
%
2) Pretraining for whole detectors. Some recent works focus on the supervised or un-supervised pre-training for instance-level visual recognition tasks among all components of detectors \cite{cai2022bigdetection,dai2021up,qi2021casp,zhao2022omdet}. Studies show that the supervised pre-training method gains more than the un-supervised one \cite{dai2021up,zhang2022dino}. However, obtaining a single large-scale object detection dataset is cost-ineffective because of the growing need for manual labeling. One substitute way is employing multiple available annotated datasets for enhancing data volume and diversity. The major challenge for unifying multiple object detection datasets is taxonomy differences. Given different datasets, the taxonomies vary widely ranging from definition and granularity even within similar semantics, which prevents us from training a universal pre-trained model.


Thus, we intend to investigate a framework that can pretrain a unified detector on multiple large-scale datasets so as to accelerate training convergence and boost performance for DETR-like detectors. We propose a framework for utilizing \textbf{M}ultiple datasets to pretrain D\textbf{ETR}-like detectors (\textbf{METR}), which can train the arbitrary number of object detection datasets jointly without the requirements for manual label taxonomy merging. We adopt the insight of divide-and-conquer to deal with object detection, \ie, first perform multi-label classification and then its byproduct is utilized for localizing the objects. Specifically, we design the Category Extraction Module (CEM) that can predict the taxonomy set contained in an image with the help of visual-language pre-trained models \cite{radford2021learning}.
Each query will be assigned one specific class from the taxonomy pool predicted by the CEM, thereby it is only responsible for predicting the instance belonging to a certain category. The detector no longer predicts the class from the whole taxonomy space, avoiding the misunderstanding upon similar semantic categories such as ``football'' and ``soccer''.
Besides, we also propose the group bipartite matching strategy to adapt our framework. It restricts the ground truth to only match queries assigned with the same category, so as to avoid the situation that the ground truth matches wrongly with the query involving similar contextual semantics but belonging to the other datasets. 


To the best of our knowledge, METR is the first DETR-style framework that is able to jointly train multiple detection datasets. 
We validate the multi-dataset learning ability of METR with four OD datasets, including COCO, Pascal VOC, Wider Face, and Wider Pedestrian. Experiments illustrate the significant performance gains over separate individual detectors. Then we explore the potential of detection pretraining by training METR on two large-scale public OD datasets (Objects365 \cite{shao2019objects365} and Openimages \cite{kuznetsova2020open}). The results on COCO show that finetuning on pre-trained models can yield a significant improvement of 4.8 mAP (47.6 vs. 52.4). In addition, our pre-trained models also show excellent transferability on DETR-like detectors for accelerating training converge and boosting performance.

In short, the contributions of this work are four folds:
\begin{itemize}
    \item We present METR, a language-aware framework for object detection pretraining utilizing multiple datasets without considering the cost of label integration. The pre-trained models can be transferred into DETR-like detectors flexibly.
    \item We propose a category extraction module to extract the potential categories in an image. Based on language embeddings, we pre-define the category for each query and enable the detector only focus on objects with specified categories.
    \item We design a group bipartite matching strategy that groups the ground truths according to the categories and restricts the ground truth to only match the query assigned to the same target category when conducting bipartite matching.
    \item Experiments show that the multi-dataset joint-trained METR outperforms the dataset-specific models on individual datasets.
    Besides, METR demonstrates a huge potential on detection pretraining in the aspects of performance and convergence, and the flexible transferability among DETR-like detectors.
\end{itemize}

\vspace{-10pt}

\begin{figure*}
\begin{center}
\centerline{\includegraphics[width=17cm]{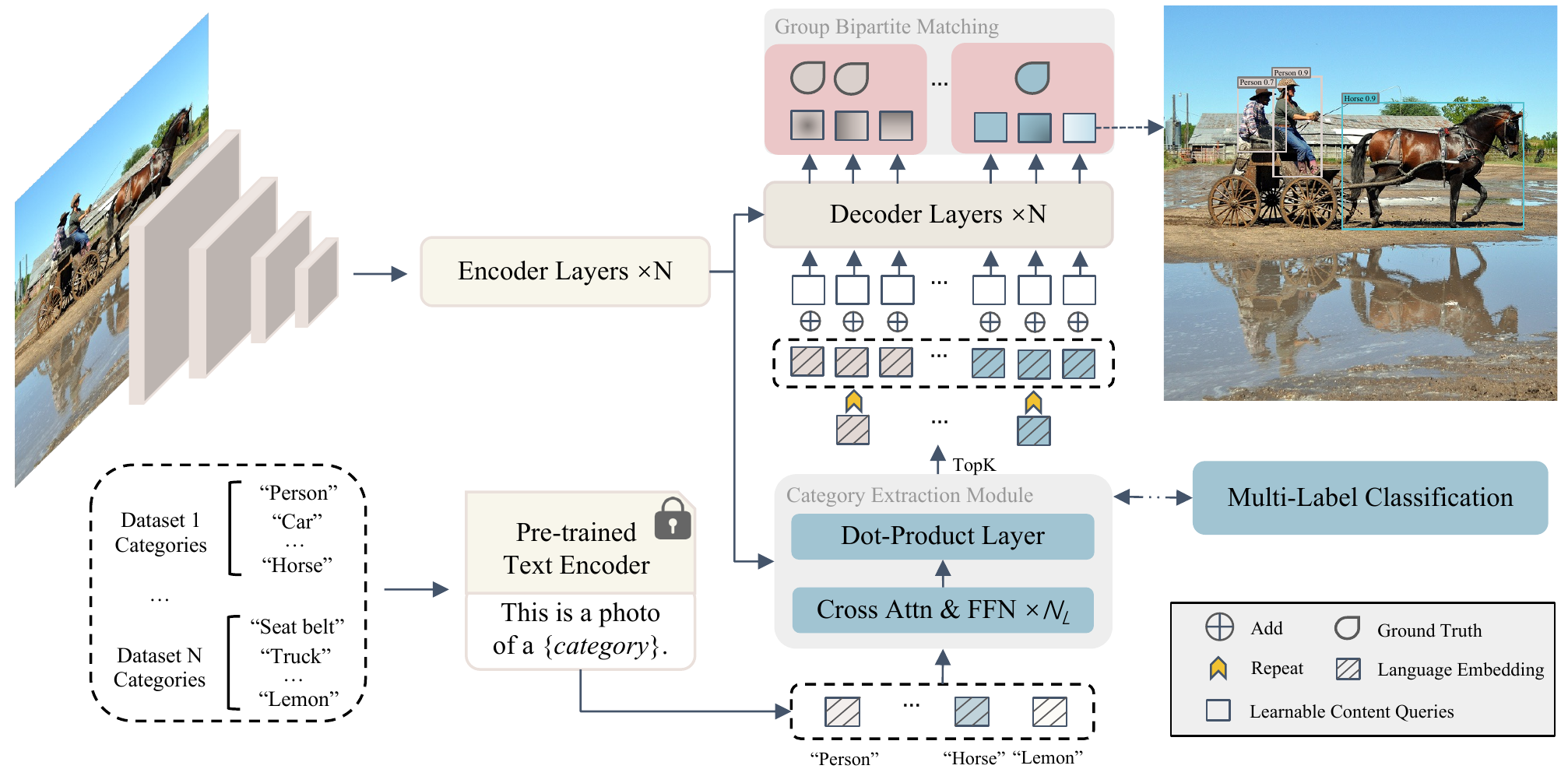}}
\end{center}
\vspace{-5pt}
   \caption{The architecture of our proposed METR. We design the category extraction module to extract the categories involved in an image and assign categories to each query with the help of language embeddings. For optimization, we change the typical bipartite matching with our group bipartite matching.}
\label{fig:pipeline}
\vspace{-4pt}
\end{figure*}

\section{Related works}

\noindent
\textbf{Multi-Dataset Object Detection (MDOD).}
Some recent works have considered using multiple datasets to jointly train a single detector to learn better feature representations. 
For example, BigDetection \cite{cai2022bigdetection} merged multiple datasets using a unified label space with a hand-designed label mapping that heavily relied on manual interference. Detection Hub \cite{meng2022detection} unified several datasets by adapting queries on category embedding based on the different distributions of datasets category. UniDet \cite{zhou2022simple} trained a unified detector with split classifiers and designed a cost of merging concepts across datasets to optimize for common taxonomy automatically. OmDet \cite{zhao2022omdet} designed a language-conditioned framework that can learn from multiple datasets, but it imposes restrictions on its own specific architecture that the pre-trained models can not transfer to other detectors. Our proposed method can conduct pretraining utilizing multiple datasets without any extra human cost for label integration.

\noindent
\textbf{Detection Pretraining.}
%
Several researchers focus on pretraining for instance-level tasks.
For example, \cite{dang2022study} designed a framework to learn spatially consistent dense representation by self-supervised pretraining, while it only achieves slight improvement during finetuning on downstream tasks.
DETReg \cite{bar2022detreg} predicted object localizations to match the localizations from an unsupervised region proposal generator and aligned the corresponding embeddings with embeddings from a frozen pre-trained image encoder. UP-DETR \cite{dai2021up} pretrained detectors by detecting random multiple query patches in an unsupervised manner. In this paper, we mainly focus on supervised pretraining that improves performance on OD tasks by making full use of available multiple annotated datasets.


\noindent
\textbf{Bipartite matching.}
The vanilla DETR \cite{carion2020end} produces a one-to-one optimal bipartite matching between predicted and ground truth objects for direct set prediction without duplicates. To enlarge the number of positive queries, Group DETR \cite{chen2021group} designed a group-wise One-to-Many assignment strategy, which adopted K groups of object queries and performed one-to-one label assignment for each group. Different from it, our proposed group bipartite matching strategy implements one-to-one matching on class-wise groups, which restricts the ground truth to only match queries assigned with the same category.
\section{Methodology}

\subsection{Overview}
The proposed framework METR mainly follows DINO \cite{zhang2022dino} which is an end-to-end object detector with well-designed strategies.
As shown in Figure \ref{fig:pipeline}, METR consists of a CNN backbone, a transformer encoder, a transformer decoder, and a Category Extraction Module (CEM). The CEM architecture is made up of several class-decoder layers and a dot-product layer \cite{radford2021learning}. The class-decoder consists of a cross-attention layer, two norm layers, and a feed-forward layer.
The detail of CEM is described in Sec. \ref{sec3.2}.
%
 Besides, we propose a group bipartite matching strategy to make one-to-one bipartite matching within class-wise groups. Sec. \ref{sec3.3} shows the details of this strategy.

\subsection{Category Extraction Module}\label{sec3.2}
The typical way to optimize a single-dataset object detector model is to use the cross-entropy loss with a fixed number of categories. For multiple datasets, the biggest challenge is label inconsistency, where taxonomy from different datasets differs, ranging from class definition and class granularity. A straightforward option to deal with this problem is manually integrating all categories into a single unified label space \cite{cai2022bigdetection,zhao2020object}. However, this could be time-consuming and error-prone because there is no explicit criterion to evaluate the quality of the unified label space. We tackle this problem by converting multiple-class prediction into binary-class prediction. The intrinsic nature of training an object detector is to classify and refine a large number of proposals \cite{tan2019learning}. 
Assuming that we possess the category set involved in an image and assign each class to multiple proposals, we can convert the proposal classification task into a binary classification task that predicts whether the proposal belongs to a specific category or not. In this case, as long as the correctness of the category set is guaranteed, the detector will not be disturbed by the label inconsistency of multiple datasets during training, thereby achieving the purpose of jointly training multiple datasets without manually integrating label space. 

\begin{figure}
\begin{center}
\includegraphics[width=0.95\linewidth]{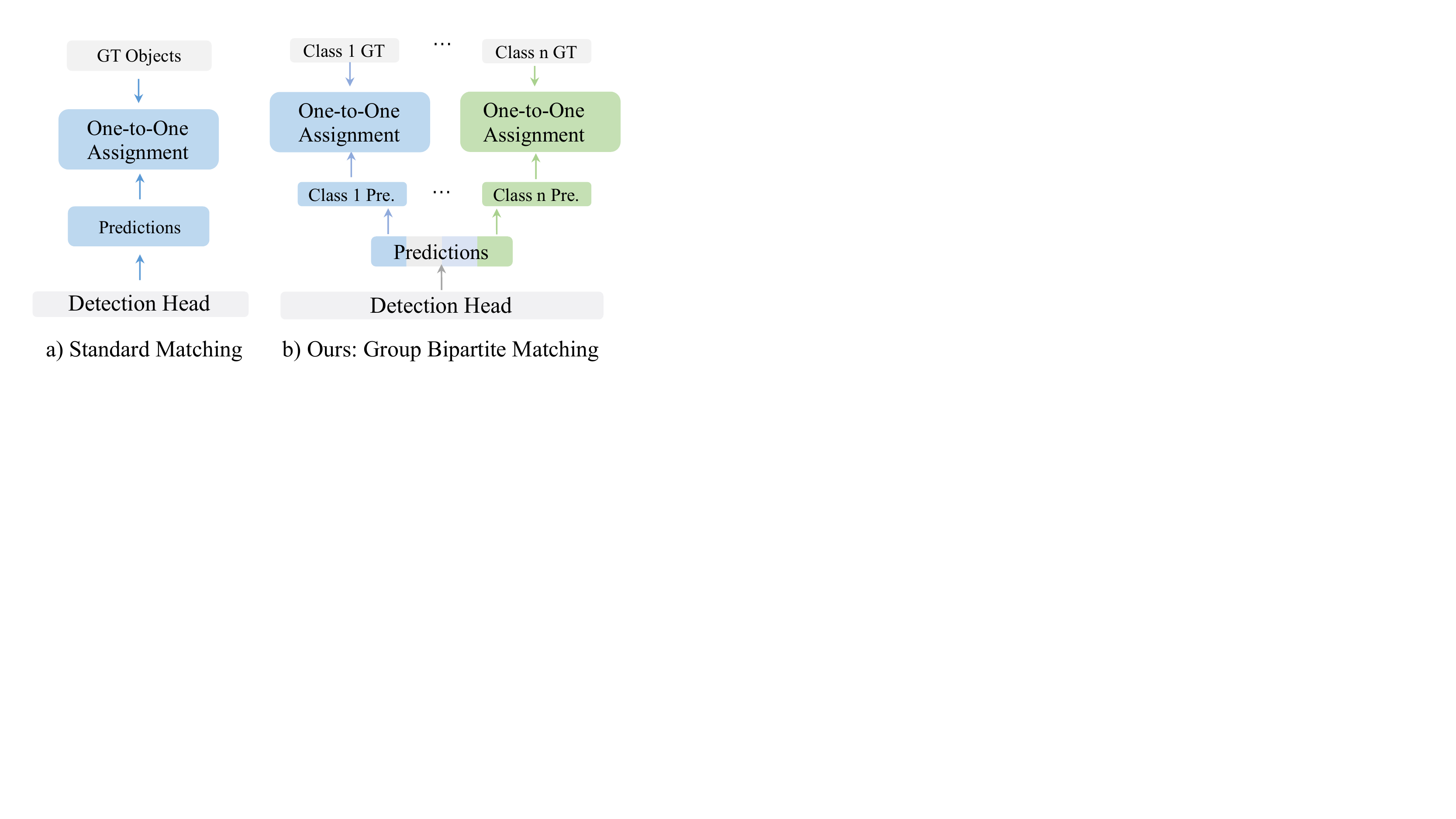}
\vspace{-17pt}
\end{center}
   \caption{Comparison with different matching strategies. a) Standard Matching: Generic matching strategy used by DETR \cite{carion2020end}. 
   b) Group Bipartite Matching: predictions are split into several parts depending on the classes, and each one is made one-to-one assignment.}
\label{fig:cagm}
\end{figure}

\vspace{-1pt}

Inspired by \cite{liu2021query2label}, we design the CEM to predict the taxonomy set involved in an image by uniting the image features and language embeddings.
Given multiple annotated datasets $\mathcal{D}=\left\{\mathcal{D}_1, \mathcal{D}_2,…, \mathcal{D}_n\right\}$ along with label space $\mathcal{Y}=\left\{\mathcal{Y}_1, \mathcal{Y}_2,…, \mathcal{Y}_n\right\}$, we take advantage of the text encoder of the CLIP \cite{radford2021learning} to generate language embedding space $\mathcal{E}=\left\{\mathcal{E}_i, | i=1,...,K \right\}$, where $K$ is the number of category in $\mathcal{Y}$. $\mathcal{E}_i \in \mathbb{R}^d$ represents the language embedding of a certain category, where $d$ is the embedding dimension size. 
%
The image features $\mathcal{F} \in \mathbb{R}^{HW \times d} $ extracted by the DETR’s encoder, where $H$ and $W$ represent the height and weight of the feature map, and language embeddings $\mathcal{E}$ are sent into CEM with cross-attention operations that can unearth the potential category set contained in images. 

As shown in Figure \ref{fig:pipeline}, CEM comprises two components, including $N_L$ class-decoder layers and a dot-product layer. The class-decoder layer is a standard Transformer architecture but removes the self-attention part. The $N_L$ class-decoder layers incrementally update the language embeddings $\mathcal{E}$ layer by layer and progressively fuse contextualized information from the input image features via cross-attention.
After that, the dot-product layer is used to predict the score $\mathcal{S}$ which indicates the existence of each category in $\mathcal{Y}$. 
We set $N_L=2$ and the whole process of CEM can be formulated as follows:
\vspace{-5pt}
\begin{align}
&	\mathcal{E}^{(1)} = FFN(CrossAttn(\mathcal{E}, \mathcal{F})), \\
&   \mathcal{E}^{(2)} = FFN(CrossAttn(\mathcal{E}^{(1)}, \mathcal{F}),\\
&   \mathcal{S} = Sigmoid(Linear(\mathcal{E}^{(2)}) \cdot \mathcal{E} / \sqrt{d} + b).
\end{align}
\vspace{-10pt}

\vspace{-5pt}
Note that $\cdot$ indicates the dot production, $\mathcal{E}^{(L)}$ is the output of the $L$-th layer of class-decoder, $d$ is the dimension of language embedding, and $b \in \mathbb{R}^K$ is a learnable bias. Then we construct the category set by selecting TopK categories with the highest scores $\mathcal{S}$, which represents that there are TopK categories contained in the certain image with the highest probability. We simply add the image-aware language embeddings $\mathcal{E}_L$ into content queries used in the detector’s decoder to pre-define the content property of each query. Each query assigned a specific category only accounts for predicting the probability of that category. In practice, we add each image-aware language embedding into multiple queries due to the fact that the instance belonging to one class may appear many times. The CEM conducts the multi-label classification task and is optimized by asymmetric loss \cite{ridnik2021asymmetric} with coefficient $\mu_{asl}$.

\vspace{-3pt}

\subsection{Group Bipartite Matching}\label{sec3.3}

We model the categories of multiple datasets into the single unified language embedding space so that we can extract the potential categories involved in an image. By pre-defining the category for each query in advance, we change the traditional multi-class prediction task by the way of measuring the matchability of pre-defined classes so as to mitigate the problem of label inconsistency among multiple datasets. However, in the training stage, the traditional bipartite matching cannot be directly applied to our method because the matching cost requires access to logists among all categories but our method can only predict the matchability for a single class. Therefore, we incorporate the group bipartite matching strategy. The comparison  with the standard matching methods \cite{carion2020end, meng2021conditional, zhudeformable, zhang2022dino} is present in Figure \ref{fig:cagm}. We first revisit the standard bipartite matching process and then describe our proposed group bipartite matching strategy. 
\vspace{-1pt}

DETR-like detectors view object detection as a set prediction problem and they rely on global One-to-One assignment for bipartite matching \cite{carion2020end}. The set prediction results contain both bounding boxes and class predictions, in which the range of class predictions is among all categories. During training, the Hungarian algorithm is employed to find an optimal global One-to-One assignment $\hat{\sigma}$ between predicted and ground truth objects:
\begin{align}
&   \hat{\sigma} = \mathop{\arg\min}_{\sigma \in \xi_N} \sum_{i = 1}^{N} \mathcal{C}_{cost} (y_i, \hat{y}_{\sigma(i)}), \\
&   \mathcal{C}_{cost} = \sum_{i = 1}^{M} [\mu_{cls} \ell_{cls}(\textbf{P}_{\hat{\sigma}(i)}, \bar{c}_i) + \ell_{box}(\textbf{b}_{\hat{\sigma}(i)}, \bar{\textbf{b}}_i)],
\end{align}
where $\xi_N$ is the set of permutations of $N$ predictions and $\mathcal{C}_{cost}$ is the matching cost between the ground truth $y_i$ and the prediction $\hat{y}_{\sigma(i)}$ with index $\sigma(i)$. Each element of ground truth consists of target class label $\bar{c}_i$ and the vector $\bar{\textbf{b}}_i$ that defines the gt box, so it can be noted as $y_i=(\bar{c}_i, \bar{\textbf{b}}_i)$. Similarity, the prediction can be defined as $\hat{y}_{\sigma(i)}=(\textbf{P}_{\hat{\sigma}(i)}, \textbf{b}_{\hat{\sigma}(i)})$. $M$ is the number of ground truth objects. $\ell_{cls}$ is the focal loss, $\mu_{cls}$ is the trade-off coefficient, and the $\ell_{box}$ is a combination of $\ell_{1}$ loss and GIoU loss. 


In order to prevent the mismatch between ground truth and the query which represents similar contextual semantics during the training stage, e.g, the ground truth is ``football'' while the query is assigned with ``soccer'', we propose the group bipartite matching strategy. It groups ground truths and queries according to their categories. Given a series of queries $Q_{\Omega}$ pre-defined class $\Omega$, we restrict the ground truth belonging to class $\Omega$ to only match with queries $Q_{\Omega}$. The group bipartite matching can be given as:
\begin{align}
&   \hat{\sigma}^\prime = \{\hat{\sigma}_g=\mathop{\arg\min}_{\sigma \in \xi_N^\prime} \sum_{i = 1}^{N^\prime} \mathcal{C}_{cost}^\prime (y_i, \hat{y}_{\sigma(i)}), | g=1,2,...,TopK\}, \\
&   \mathcal{C}_{cost}^\prime = \sum_{i = 1}^{M^\prime} [\mu_{cls} \ell_{match}({\textbf{P}_{\hat{\sigma}(i)}}^\prime, {\bar{c}_i}^\prime) + \ell_{box}(\textbf{b}_{\hat{\sigma}(i)}, \bar{\textbf{b}}_i)], 
\end{align}
where $N^\prime$ is the number of queries that account for predicting class $\Omega$, and $M^\prime$ is the number of ground truth belonging to the class $\Omega$. The difference between Eq(5) and Eq(7) is that we replace the $\ell_{cls}$ with $\ell_{match}$. $\ell_{match}$ is simply implemented by a binary focal loss. ${\textbf{P}_{\hat{\sigma}(i)}}^\prime$ is the matchability of the class $\Omega$ and ${\bar{c}_i}^\prime=1$ which indicates this is a ground truth of class $\Omega$. The training loss is the same as the $\mathcal{C}_{cost}^\prime$ except that only performed on matched pairs. 

\vspace{-2pt}

\section{Experiments}


We first present our experimental setup in Sec. \ref{sec:4.1}, and then present competitive results on different tasks in Sec. \ref{sec:4.2}, including multi-dataset joint training, pretraining \& finetune paradigm, \etc. 
Finally, we conduct thorough ablation studies to better understand how each part affects the performance in Sec. \ref{sec:4.3}.

\begin{table*}[!t]
\centering
\caption{Results of MDOD on four datasets. Compared with the well-designed OmDet \cite{zhao2022omdet}, METR achieves higher performance. ``Single'' means training detector on the four datasets separately, and ``Joint'' means using all datasets for training.}

\begin{tabular}{lcccccccccccc}
\toprule
     & \multicolumn{3}{c}{COCO} & \multicolumn{3}{c}{PASCAL VOC} & \multicolumn{3}{c}{WIDER FACE} & \multicolumn{3}{c}{WIDER Pedestrain}   \\
     & mAP    & AP50    & AP75   & mAP    & AP50  & AP75   & mAP  & AP50  & AP75   & mAP   & AP50   & AP75    \\ \midrule
OmDet-Single   & 43.0  & 62.4  & 46.7  & 45.5  & 65.9  & 49.4  & 23.4 & 44.8 & 21.9 & 33.9 & 55.2 & 37.0 \\
OmDet-Joint & 42.2   & 61.4  & 54.6  & 60.8  & 82.1  & 66.6  & 30.7  & 57.2 & 30.0 & 46.8 & 74.0 & 50.9 \\ \midrule
METR-Single    & \textbf{47.6}  & \textbf{65.6}  & \textbf{52.0}  & 55.9  & 77.0  & 60.4  & 28.5 & 55.7 & 26.6 & 57.3 & 84.1 & 66.7 \\
METR-Joint & 46.6  & 64.2  & 50.5  & \textbf{65.1}  & \textbf{84.9}  & \textbf{71.7}  & \textbf{31.8} & \textbf{58.6} & \textbf{31.3} & \textbf{62.8} & \textbf{89.2} & \textbf{73.5}  \\ 
\bottomrule    
\end{tabular}
\label{table:mdod}
\end{table*}


\begin{table*}[!t]
\begin{center}
\caption{Comparsion with METR and other DETRs on COCO val2017. ``Defor'' denotes the abbreviation of ``Deformable''.
``DC5'' means using dilated larger resolution feature map and the others use standard 4-scale features. The ``IN1K'', ``O365'', and ``OImg'' refer to the ``ImageNet-1K'', ``Objects365'', and ``OpenImages'', respectively. }
\begin{tabular}{p{3.7cm}p{0.7cm}<{\centering}p{0.7cm}<{\centering}p{1.5cm}<{\centering}p{0.8cm}<{\centering}p{0.9cm}<{\centering}p{1.2cm}<{\centering}p{0.6cm}<{\centering}p{0.6cm}<{\centering}p{0.6cm}<{\centering}p{0.6cm}<{\centering}p{0.6cm}<{\centering}}
\toprule
Model                       & \#Query & \#Epoch & Pretrain & Params & GFLOPS & mAP     & AP50 & AP75 & APs  & APm  & APl  \\
\midrule
DETR(DC5)                   & 100     & 500   & IN1K   & 41M    & 187    & 43.3    & 63.1  & 45.9  & 22.5 & 47.3 & 61.1 \\
Conditional DETR(DC5)       & 300     & 50    & IN1K   & 44M    & 195    & 43.8    & 64.4  & 46.7  & 24.0 & 47.6 & 60.7 \\
DAB-DETR(DC5)               & 900     & 50    & IN1K   & 44M    & 216    & 45.7    & 66.2  & 49.0  & 26.1 & 49.4 & 63.1 \\
SAP-DETR(DC5)               & 300     & 50    & IN1K   & 47M    & 197    & 46.0    & 65.5  & 48.9  & 26.4 & 50.2 & 62.6 \\
Defor DETR            & 300     & 50   & IN1K    & 40M    & 173    & 43.8    & 62.6  & 47.7  & 26.4 & 47.1 & 58.0 \\ 
DAB-Defor-DETR  & 300     & 50    & IN1K   & 48M  & 195    & 48.7    & 67.2  & 53.0  & 31.4 & 51.6 & 63.9 \\ \midrule
DN-Defor-DETR & 900     & 12    & IN1K   & 48M    & 195    & 43.4    & 61.9  & 47.2  & 24.8 & 46.8 & 59.4 \\
DINO               & 100     & 12   & IN1K    & 47M    & 227    & 47.2    & 64.5  & 51.3  & 29.4 & 50.5 & 61.9 \\
METR               & 100     & 12   & IN1K    & 50M    & 231    & \textbf{47.6(+0.4)}    & 65.6  & 52.0  & 29.8 & 50.9 & 62.6 \\ \midrule
DINO               & 100     & 50   & IN1K    & 47M    & 227    & 49.3    & 67.6  & 53.5  & 33.0 & 52.6 & 63.6 \\
METR              & 100     & 50   & IN1K    & 50M    & 231    & \textbf{49.7(+0.4)}    & 67.6  & 53.8  & 32.2 & 52.6  & 64.9 \\  \midrule
DINO               & 100     & 12   & O365    & 47M    & 227    & 50.7    & 68.0  & 55.4  & 35.0 & 54.2 & 64.2 \\
METR               & 100     & 12   & O365+OImg    & 50M    & 231    & \textbf{52.4(+1.7)}    & 70.3  & 57.5  & 35.7 & 55.9 & 66.4 \\ 
\bottomrule
\end{tabular}
\label{table:single_dataset}
\vspace{-8pt}
\end{center}
\end{table*}

\subsection{Experimental Setup}
\label{sec:4.1}
\noindent
\textbf{Datasets.}
We perform multiple datasets jointly training and pre-training \& finetuning to verify the performance of METR.
For multiple datasets jointly training, we follow the experiments setting from \cite{zhao2022omdet} and choose COCO, Pascal VOC, WIDER FACE \cite{yang2016wider}, and WIDER Pedestrian \cite{loy2019wider} as joint training datasets. As for pre-training METR, we select two large-scale public object detection datasets including OpenImages V4 \cite{kuznetsova2020open} and Object365 \cite{shao2019objects365}. OpenImages has 14.6M bounding boxes upon 1.74M images over 600 object classes. Note that we replace the official noisy annotation that contains 10\% labels generated semi-automatically using \cite{papadopoulos2016we} with the one released by BigDetection \cite{cai2022bigdetection}.
Object365 is another large-scale dataset, and it contains around 1.72M images with more than 22.8M bounding boxes over 365 categories. Then we finetune the pre-trained models on the COCO 2017 dataset.
%

\noindent
\textbf{Implementation details.}
The classification head of METR is the form of the dot-product layer \cite{radford2021learning} and the detailed algorithm will be summarized in supplementary materials. We conduct our experiments using the PyTorch~\cite{paszke2019pytorch} deep learning framework. We use ResNet50 \cite{he2016deep} as the backbone with an ImageNet-pretrained model from TORCHVISION in all experiments unless specified otherwise. The AdamW \cite{loshchilov2017decoupled} optimizer is used. The learning rates for the backbone and the transformer are initially set to be $1\mathrm{e}{-5}$ and $1\mathrm{e}{-4}$, respectively. The learning rate is dropped by a factor of 10 after 11 epochs for 12 training epochs, and after 40 epochs for 50 training epochs. The weight decay is set to be $1\mathrm{e}{-4}$. We train METR on COCO using 8 Nvidia A100 40G GPUs, and each GPU has a local batch size of 1 image only. For language embeddings, we select CLIP-B/16 \cite{radford2021learning} text encoder throughout this study. We adopt most of the default hyper-parameters and data augmentation same as DINO. 
\begin{table*}[]
\centering
\caption{The transferability of our pre-trained METR. The pretrain column with marked \ding{55} denotes loading ImageNet-1K pretrained backbone and \checkmark denotes loading the pre-trained METR model. 
Note that with pre-trained METR, the weights of the backbone, encoder, and decoder can be loaded.}
\begin{tabular}{p{4.2cm}p{0.8cm}<{\centering}p{0.8cm}<{\centering}p{0.8cm}<{\centering}p{1.2cm}<{\centering}p{0.6cm}<{\centering}p{0.6cm}<{\centering}p{1.2cm}<{\centering}p{0.6cm}<{\centering}p{0.6cm}<{\centering}}
\toprule
Model            & \#Query & Pretrain & \#Epoch & mAP      & AP50 & AP75 & APs  & APm  & APl  \\
\midrule
Deformable DETR  & 300     & \ding{55}        & 50      & 43.5    & 62.5 & 47.3 & 26.2 & 46.9 & 57.7     \\
Deformable DETR & 300     & \checkmark       & 50      & \textbf{46.2(+2.7)}    & 65.1 & 50.8 & \textbf{28.6(+2.4)} & 50.1 & 60.4 \\
\midrule
DAB-Deformable DETR  & 300     & \ding{55}        & 50      & 48.7    & 67.2 & 53.0 & 31.4 & 51.6 & 63.9     \\
DAB-Deformable DETR  & 300     & \checkmark       & 50      & \textbf{51.3(+2.6)} & 69.1 & 55.9 & \textbf{33.9(+2.5)} & 54.5 & 65.5 \\
\midrule
DN-DAB-Deformable DETR  & 300     & \ding{55}        & 50      & 49.5    & 67.6 & 53.8 & 31.3 & 52.6 & 65.4     \\
DN-DAB-Deformable DETR  & 300     & \checkmark       & 50      & \textbf{51.3(+1.8)}    & 69.2 & 55.6 & \textbf{34.1(+3.8)} & 54.8 & 66.1 \\
\midrule
DINO             & 100     & \ding{55}        & 12      & 47.2    & 64.5 & 51.3 & 29.4 & 50.5 & 61.9 \\
DINO             & 100     & \checkmark       & 12      & \textbf{50.6(+3.4)}    & 67.5 & 55.3 & \textbf{33.5(+4.6)} & 54.0 & 64.4 \\
\bottomrule
\end{tabular}
\label{tab:transfer}
\end{table*}
\subsection{Comparison with Existing Methods}
\label{sec:4.2}

\noindent
\textbf{Detection on Multi-Dataset.}
For MDOD, we select four datasets to conduct joint training experiments with the same setting as OmDet \cite{zhao2022omdet}. We compare the performance with OmDet, which is also a detection framework enabling joint training. OmDet uses ImageNet pre-trained Swin Transformer Tiny \cite{liu2021swin} as the backbone while METR uses ImageNet pre-trained ResNet50 backbone, the parameters of these two backbones are comparable (29M vs. 26M). 
The results are shown in Table \ref{table:mdod}. 

As for the performance of training on individual datasets, METR-Single outperforms OmDet-Single by a large margin, which verifies that METR is an extremely strong and general object detector. In addition, the mAP of METR-Joint is significantly higher than METR-Single on PASCAL VOC (+9.2 AP), WIDER FACE (+3.3 AP), and WIDER Pedestrian (+5.5 AP). 
Because of the knowledge sharing across several datasets, the performance of joint training performs better than isolated individual ones.
These results confirm that METR does not suffer from taxonomy conflicts. However, we find a similar situation with OmDet in that the mAP on COCO slightly drops in the joint training stage. We suspect the main reason for the slight drop is that many objects belonging to the categories of COCO that exist in the other three datasets are considered backgrounds. 

\noindent
\textbf{Comparaison on DETR-like detectors.}
We also explore the performance of METR on pretrain \& finetune paradigm. Table \ref{table:single_dataset} shows the results on the COCO val2017. We compare METR with several strong DETR-like methods \cite{carion2020end,li2022dn,liu2022dab,liu2022sap,meng2021conditional,zhang2022dino}. For fairness, we also report the number of parameters and GFLOPS.
%
%
As shown in Table \ref{table:single_dataset}, METR shows effectiveness in improving both convergence speed and performance, that is, with 100 queries and 12-epoch setting, METR outperforms most DETR-like detectors which use more queries and train longer epochs.
For example, DAB-DETR achieves 45.7 mAP with 900 queries and 50-epoch setting, while METR achieves 47.6 mAP.
In the condition of 12 epochs and 50 epochs with ImageNet1K pre-trained model, METR surpasses DINO by 0.4 mAP. 
%

Furthermore, we probe the effectiveness of pretrain \& finetune paradigm. Considering that DINO cannot train multi-dataset jointly, thus we pretrain it on Objects365 because the Objects365 dataset is a high-quality fully-annotated dataset that contains the all categories defined in COCO. 
 We use two publicly available large-scale OD datasets including Openimages V4 and Objects365 for pretraining METR. The experiment setting is pretraining for 4 epochs and finetuning for 12 epochs considering the training time cost.
 After changing the pre-trained datasets from ImagetNet1k to Objects365 or ``Objects365 \& OpenImages'', the improvement of METR compared with DINO increases from +0.4 mAP (47.2 $\to$ 47.6) to +1.7 mAP (50.7 $\to$ 52.4). This demonstrates the huge potential for pretraining among multi-dataset jointly and we think that as the number of pre-trained datasets and training epochs increases, the improvements will be higher in the finetuning stage.

\noindent
\textbf{Transferability.}
We verify the transferability of our pre-trained model, which is pretrained on Objects365 and OpenImages jointly for 4 epochs. Table \ref{tab:transfer} reports the performance comparison for whether loading METR pre-trained models on multiple DETR-like detection frameworks.
Note that all methods use ResNet50 as the backbone and deformable attention \cite{zhudeformable} in the transformer encoder and decoder.
Therefore, we can load our pre-trained model for all Deformable-DETR families' methods with their backbone, encoder, and decoder.
As for all those methods, loading METR pre-trained models can bring large performance gains.
For example, under the 50-epoch and 300-query setting, Deformable-DETR, DAB-Deformable-DETR, and DN-DAB-Deformable DETR loading our pre-trained model can outperform the baseline by 2.7 mAP, 2.6 mAP, and 1.8 mAP, respectively. Besides, the experiment on DINO shows that the METR pre-trained model can obtain non-trivial improvement (+3.2 mAP) over the baseline. All of these gains prove that our METR pre-trained model has strong and flexible transferability. In addition, we are surprised to discover that our METR pre-trained model greatly promotes performance on small objects. We reckon that rich semantical representations of pre-trained models can make detectors recognize small objects easier. 

\begin{figure*}
\centerline{\includegraphics[width=0.95\textwidth]{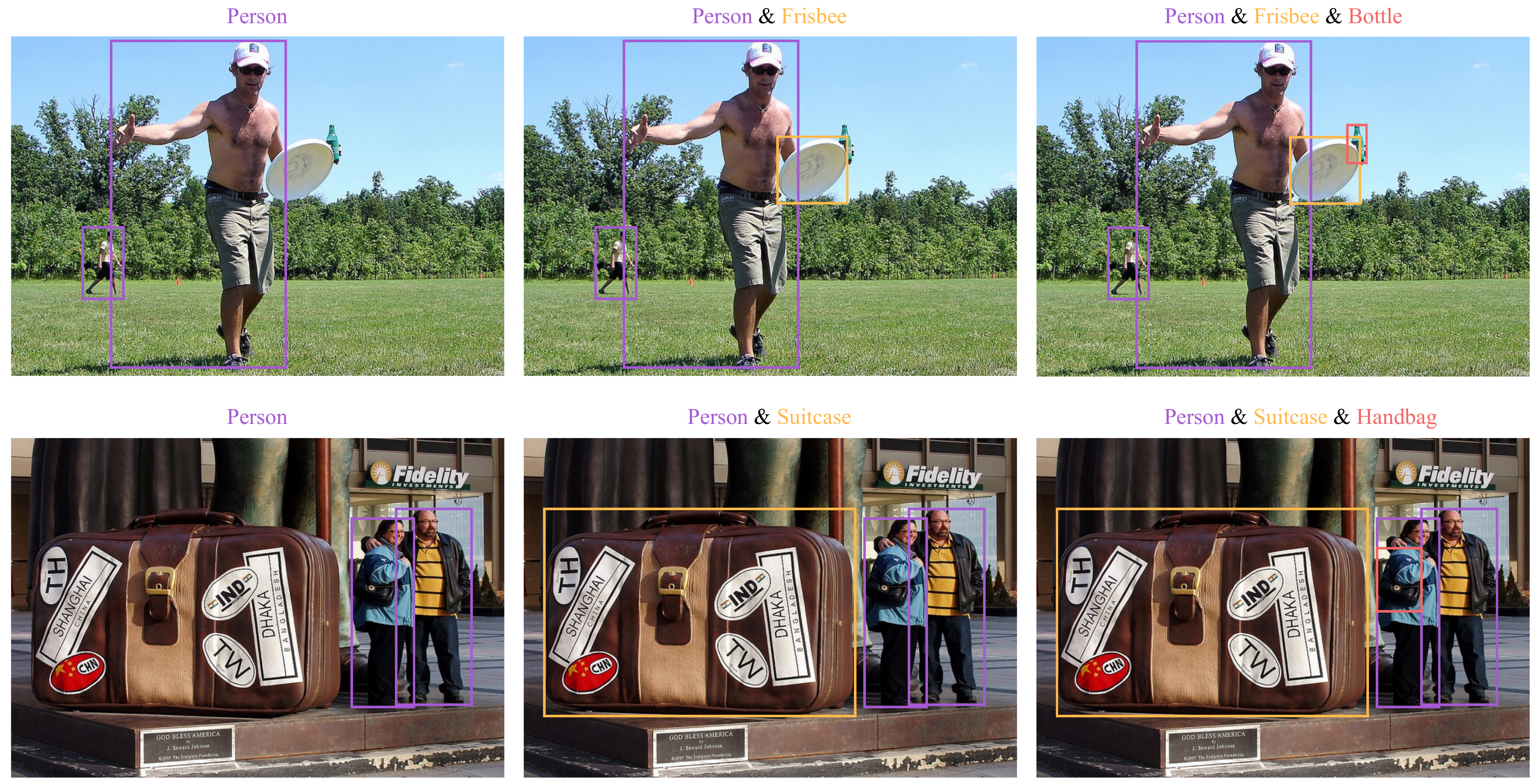}}
\caption{Visualization of Language-Aware Detection.
With different input texts, METR can dynamically modify its object localization and classification results.
For a clear illustration, instead of keeping a mass of high-score results in the eval process, we only keep the predicted bounding boxes with scores larger than 0.3.}
\label{fig:Language-Aware-Detection}
\vspace{-5pt}
\end{figure*}

\subsection{Ablation studies.}
\label{sec:4.3}
\noindent
\textbf{Importance of CEM.}
 We find it feasible that train METR without our proposed CEM because CEM aims to discover the potential categories involved in images while we certainly can pre-obtain the categories from the ground truth. While for inference, we have to predict objects among all categories due to the lack of information on categories included in images. Table \ref{table:ablation_study} reveals the effectiveness of CEM. There is a 4.6 mAP drop when we discard the CEM, thus indicating the effectiveness of our proposed CEM. As for the language embeddings generated by CLIP models should be learnable or fixed, we observe that it leads 1.0 mAP decrease when fixing language embeddings. We suspect that embedding generated by CLIP may not be well aligned with image features. It needs to be tuned into the training process. Moreover, CEM consists of two layers of class-decoder and introduces extra 3M parameters on top of DINO. Thus, we reduce the METR’s decoder layers from 6 to 4 for a fair parameter comparison. The performance slightly drops from 47.6 to 47.2, which demonstrates that our method is less influential on the number of decoder layers. Under the comparable parameter setting, the mAP of METR and DINO is equal but the AP50 of METR is higher 0.5 than DINO (65.0 vs. 64.5), which shows that METR is a strong and general object detector. Besides, our proposed CEM is able to perform multi-label classification and we will elaborate on its detailed results in supplementary materials.
\begin{table}[!t]
\centering
\caption{Ablation comparison of the proposed CEM. 
CEM consists of two class-decoder layers.
\dag means the language embedding is unlearnable.}
\begin{tabular}{p{1.1cm}p{0.6cm}<{\centering}p{0.6cm}<{\centering}p{0.6cm}<{\centering}p{0.8cm}<{\centering}p{0.7cm}<{\centering}p{0.7cm}<{\centering}}
\toprule
Model & \#TopK & \#Dec & CEM & Params    & mAP  & mAP50 \\
\midrule
METR     & 20     & 6     & \checkmark     & 49.5M & 47.6  & 65.5 \\
METR     & 20     & 6     & -     & 46.8M & 43.0  & 59.1 \\ 
METR\dag & 20     & 6     & \checkmark     & 49.5M & 46.6  & 64.0 \\ 
\midrule
DINO     & -      & 6     & -     & 46.5M & 47.2  & 64.5 \\
METR     & 20     & 4     & \checkmark     & 46.5M & 47.2  & 65.0 \\ 
\bottomrule
\end{tabular}
\vspace{-5pt}
\label{table:ablation_study}
\end{table}
\begin{table}[!t]
\centering
\caption{Results of large-scale datasets pretraining.
``*'' denotes employing a layer-decay strategy in the process of finetuning. We use the terms “OImg” and “O365” to denote the Openimages v4 and Objects365 datasets, respectively.}
\begin{tabular}{lclccccccccc}
\toprule
Model & \#query & Pretrain                     & mAP   & AP50 \\
\midrule
METR  & 100     & None                        & 47.6 & 65.5  \\
METR  & 100     & OImg           & \textbf{49.5(+1.9)} & 67.4 &   \\
METR  & 100     & OImg+O365      & \textbf{51.7(+2.2)}     & 69.7    \\
METR*  & 100     & OImg+O365     & \textbf{52.4(+0.7)}     & 70.3   \\
\bottomrule
\end{tabular}
\vspace{-5pt}
\label{table:pretraining}
\end{table}

\noindent
\textbf{The number of datasets for pretraining.}
Table \ref{table:pretraining} shows the improvement by using two datasets for pretraining. First, we pretrain METR on Openimages V4 and then finetune it on COCO. After loading the pre-trained model on Openimages V4, METR achieves an improvement of +1.9 mAP. Then we pretrain METR using both Openimages V4 and Object365 datasets. Adding Object365 into pretraining datasets leads to a 2.2 AP improvement (49.5 $\to$ 51.7) on COCO validation.
Following \cite{chen2022group}, we apply a layer-decay strategy (see details in supplementary materials) during finetuning and it gives 0.7 mAP gain. 

\begin{table}[!t]
\centering
\caption{The effect of two hyper-parameters.}
\begin{tabular}{cccccc}
\toprule
TopK & 10 & 15 & 20   & 25   & 30   \\
\midrule
mAP  & 47.2 & 47.2 & 47.6 & 47.4 & 46.9 \\
\midrule
$\mu_{asl}$ & 0.2 & 0.5 & 1   & 2   & 5   \\
\midrule
mAP  & 47.5 & 47.6 & 47.6 & 46.1 & 44.1 \\
\bottomrule
\end{tabular}
\vspace{-5pt}
\label{table:TopK Ablation}
\end{table}
\noindent
\textbf{Hyper-parameters.}
In Table \ref{table:TopK Ablation}, we show the effect of using different hyper-parameters on TopK and $\mu_{asl}$.
%
First, TopK is relevant to the maximum number of categories contained in one image. We make a statistic on the COCO dataset and the maximum number of categories in one sample is 18. Hence the optimal parameter of TopK is close to 18. When TopK=10, the model is unable to predict all foregrounds so there is a slight degradation from 47.6 to 47.2. As for TopK=30, the performance drops 0.7 mAP due to the raising of difficulty for model optimization.
Then, CEM uses asymmetric loss (ASL) for the multi-label classification task. We observe that the $\mu_{asl}$ is relatively robust in the range from 0.5 to 1. All experiments set $\mu_{asl}=1$.

\subsection{Language-Aware Detection}
To figure out the language embeddings whether can specify the queries' semantic representation, we visualize the predictions among single language embedding and the combination of multiple language embeddings. Figure \ref{fig:Language-Aware-Detection} shows the visualization of language-aware detection by combining different classes.
For a clear illustration, we only keep the predicted bounding boxes with scores greater than 0.3.
First, input with only a single text, \eg \textit{[person]}, METR can detect the corresponding instances correctly.
Then, inputting different combined texts, \eg \textit{[Person, Frisbee]} and \textit{[Person, Frisbee, Bottle]}, METR can dynamically adapt
its object detection results conditioned on the given texts. To some extent, this characteristic also illustrates that the language embeddings indeed have the capacity of assigning contextual class information for queries. 
%

\section{Conclusion}

We present the METR which solves the taxonomy conflict problem during multi-dataset joint training and explore the huge potential of object detection pretraining. It pre-defines the categories for each query so that converting the typical multi-classification in object detection into binary classification. The proposed category extraction module and group bipartite matching strategy help us implement our core idea to adapt the DETR-like framework. Experiments show that METR performs well on multi-task joint training and our pre-trained models have high flexible transferability and boost the performance upon various DETR-like detectors. Future research will focus on exploiting the infinite images on the Web to conduct semi-supervised or unsupervised object detection pretraining. 

{\small
\bibliographystyle{ieee_fullname}
\bibliography{egbib}

\begin{thebibliography}{10}\itemsep=-1pt

\bibitem{arnab2021vivit}
Anurag Arnab, Mostafa Dehghani, Georg Heigold, Chen Sun, Mario Lu{\v{c}}i{\'c},
  and Cordelia Schmid.
\newblock Vivit: A video vision transformer.
\newblock In {\em ICCV}, pages 6836--6846, 2021.

\bibitem{bar2022detreg}
Amir Bar, Xin Wang, Vadim Kantorov, Colorado~J Reed, Roei Herzig, Gal Chechik,
  Anna Rohrbach, Trevor Darrell, and Amir Globerson.
\newblock Detreg: Unsupervised pretraining with region priors for object
  detection.
\newblock In {\em CVPR}, pages 14605--14615, 2022.

\bibitem{cai2022bigdetection}
Likun Cai, Zhi Zhang, Yi Zhu, Li Zhang, Mu Li, and Xiangyang Xue.
\newblock Bigdetection: A large-scale benchmark for improved object detector
  pre-training.
\newblock In {\em CVPR}, pages 4777--4787, 2022.

\bibitem{cai2018cascade}
Zhaowei Cai and Nuno Vasconcelos.
\newblock Cascade r-cnn: Delving into high quality object detection.
\newblock In {\em CVPR}, pages 6154--6162, 2018.

\bibitem{carion2020end}
Nicolas Carion, Francisco Massa, Gabriel Synnaeve, Nicolas Usunier, Alexander
  Kirillov, and Sergey Zagoruyko.
\newblock End-to-end object detection with transformers.
\newblock In {\em ECCV}, pages 213--229, 2020.

\bibitem{chen2021group}
Qiang Chen, Xiaokang Chen, Jian Wang, Haocheng Feng, Junyu Han, Errui Ding,
  Gang Zeng, and Jingdong Wang.
\newblock Group detr: Fast detr training with group-wise one-to-many
  assignment.
\newblock {\em arXiv preprint arXiv:2207.13085}, 1(2), 2022.

\bibitem{chen2022group}
Qiang Chen, Jian Wang, Chuchu Han, Shan Zhang, Zexian Li, Xiaokang Chen, Jiahui
  Chen, Xiaodi Wang, Shuming Han, Gang Zhang, et~al.
\newblock Group detr v2: Strong object detector with encoder-decoder
  pretraining.
\newblock {\em arXiv preprint arXiv:2211.03594}, 2022.

\bibitem{chen2019med3d}
Sihong Chen, Kai Ma, and Yefeng Zheng.
\newblock Med3d: Transfer learning for 3d medical image analysis.
\newblock {\em arXiv preprint arXiv:1904.00625}, 2019.

\bibitem{chen2022context}
Xiaokang Chen, Mingyu Ding, Xiaodi Wang, Ying Xin, Shentong Mo, Yunhao Wang,
  Shumin Han, Ping Luo, Gang Zeng, and Jingdong Wang.
\newblock Context autoencoder for self-supervised representation learning.
\newblock {\em arXiv preprint arXiv:2202.03026}, 2022.

\bibitem{dai2021dynamic}
Xiyang Dai, Yinpeng Chen, Jianwei Yang, Pengchuan Zhang, Lu Yuan, and Lei
  Zhang.
\newblock Dynamic detr: End-to-end object detection with dynamic attention.
\newblock In {\em ICCV}, pages 2988--2997, 2021.

\bibitem{dai2021up}
Zhigang Dai, Bolun Cai, Yugeng Lin, and Junying Chen.
\newblock Up-detr: Unsupervised pre-training for object detection with
  transformers.
\newblock In {\em CVPR}, pages 1601--1610, 2021.

\bibitem{dang2022study}
Trung Dang, Simon Kornblith, Huy~Thong Nguyen, Peter Chin, and Maryam Khademi.
\newblock A study on self-supervised object detection pretraining.
\newblock {\em arXiv preprint arXiv:2207.04186}, 2022.

\bibitem{ge2021yolox}
Zheng Ge, Songtao Liu, Feng Wang, Zeming Li, and Jian Sun.
\newblock Yolox: Exceeding yolo series in 2021.
\newblock {\em arXiv preprint arXiv:2107.08430}, 2021.

\bibitem{he2022masked}
Kaiming He, Xinlei Chen, Saining Xie, Yanghao Li, Piotr Doll{\'a}r, and Ross
  Girshick.
\newblock Masked autoencoders are scalable vision learners.
\newblock In {\em CVPR}, pages 16000--16009, 2022.

\bibitem{he2020momentum}
Kaiming He, Haoqi Fan, Yuxin Wu, Saining Xie, and Ross Girshick.
\newblock Momentum contrast for unsupervised visual representation learning.
\newblock In {\em CVPR}, pages 9729--9738, 2020.

\bibitem{he2016deep}
Kaiming He, Xiangyu Zhang, Shaoqing Ren, and Jian Sun.
\newblock Deep residual learning for image recognition.
\newblock In {\em CVPR}, pages 770--778, 2016.

\bibitem{kuznetsova2020open}
Alina Kuznetsova, Hassan Rom, Neil Alldrin, Jasper Uijlings, Ivan Krasin, Jordi
  Pont-Tuset, Shahab Kamali, Stefan Popov, Matteo Malloci, Alexander
  Kolesnikov, et~al.
\newblock The open images dataset v4: Unified image classification, object
  detection, and visual relationship detection at scale.
\newblock {\em International Journal of Computer Vision (IJCV)},
  128(7):1956--1981, 2020.

\bibitem{li2022dn}
Feng Li, Hao Zhang, Shilong Liu, Jian Guo, Lionel~M Ni, and Lei Zhang.
\newblock Dn-detr: Accelerate detr training by introducing query denoising.
\newblock In {\em CVPR}, pages 13619--13627, 2022.

\bibitem{liu2022dab}
Shilong Liu, Feng Li, Hao Zhang, Xiao Yang, Xianbiao Qi, Hang Su, Jun Zhu, and
  Lei Zhang.
\newblock Dab-detr: Dynamic anchor boxes are better queries for detr.
\newblock {\em arXiv preprint arXiv:2201.12329}, 2022.

\bibitem{liu2021query2label}
Shilong Liu, Lei Zhang, Xiao Yang, Hang Su, and Jun Zhu.
\newblock Query2label: A simple transformer way to multi-label classification.
\newblock {\em arXiv preprint arXiv:2107.10834}, 2021.

\bibitem{liu2022sap}
Yang Liu, Yao Zhang, Yixin Wang, Yang Zhang, Jiang Tian, Zhongchao Shi,
  Jianping Fan, and Zhiqiang He.
\newblock Sap-detr: Bridging the gap between salient points and queries-based
  transformer detector for fast model convergency.
\newblock {\em arXiv preprint arXiv:2211.02006}, 2022.

\bibitem{liu2021swin}
Ze Liu, Yutong Lin, Yue Cao, Han Hu, Yixuan Wei, Zheng Zhang, Stephen Lin, and
  Baining Guo.
\newblock Swin transformer: Hierarchical vision transformer using shifted
  windows.
\newblock In {\em ICCV}, pages 10012--10022, 2021.

\bibitem{loshchilov2017decoupled}
Ilya Loshchilov and Frank Hutter.
\newblock Decoupled weight decay regularization.
\newblock {\em arXiv preprint arXiv:1711.05101}, 2017.

\bibitem{loy2019wider}
Chen~Change Loy, Dahua Lin, Wanli Ouyang, Yuanjun Xiong, Shuo Yang, Qingqiu
  Huang, Dongzhan Zhou, Wei Xia, Quanquan Li, Ping Luo, et~al.
\newblock Wider face and pedestrian challenge 2018: Methods and results.
\newblock {\em arXiv preprint arXiv:1902.06854}, 2019.

\bibitem{meng2021conditional}
Depu Meng, Xiaokang Chen, Zejia Fan, Gang Zeng, Houqiang Li, Yuhui Yuan, Lei
  Sun, and Jingdong Wang.
\newblock Conditional detr for fast training convergence.
\newblock In {\em ICCV}, pages 3651--3660, 2021.

\bibitem{meng2022detection}
Lingchen Meng, Xiyang Dai, Yinpeng Chen, Pengchuan Zhang, Dongdong Chen,
  Mengchen Liu, Jianfeng Wang, Zuxuan Wu, Lu Yuan, and Yu-Gang Jiang.
\newblock Detection hub: Unifying object detection datasets via query
  adaptation on language embedding.
\newblock {\em arXiv preprint arXiv:2206.03484}, 2022.

\bibitem{papadopoulos2016we}
Dim~P Papadopoulos, Jasper~RR Uijlings, Frank Keller, and Vittorio Ferrari.
\newblock We don't need no bounding-boxes: Training object class detectors
  using only human verification.
\newblock In {\em CVPR}, pages 854--863, 2016.

\bibitem{paszke2019pytorch}
Adam Paszke, Sam Gross, Francisco Massa, Adam Lerer, James Bradbury, Gregory
  Chanan, Trevor Killeen, Zeming Lin, Natalia Gimelshein, Luca Antiga, et~al.
\newblock Pytorch: An imperative style, high-performance deep learning library.
\newblock In {\em NIPS}, 2019.

\bibitem{qi2021casp}
Lu Qi, Jason Kuen, Zhe Lin, Jiuxiang Gu, Fengyun Rao, Dian Li, Weidong Guo,
  Zhen Wen, and Jiaya Jia.
\newblock Casp: Class-agnostic semi-supervised pretraining for detection and
  segmentation.
\newblock {\em arXiv preprint arXiv:2112.04966}, 2021.

\bibitem{radford2021learning}
Alec Radford, Jong~Wook Kim, Chris Hallacy, Aditya Ramesh, Gabriel Goh,
  Sandhini Agarwal, Girish Sastry, Amanda Askell, Pamela Mishkin, Jack Clark,
  et~al.
\newblock Learning transferable visual models from natural language
  supervision.
\newblock In {\em ICML}, pages 8748--8763, 2021.

\bibitem{ridnik2021asymmetric}
Tal Ridnik, Emanuel Ben-Baruch, Nadav Zamir, Asaf Noy, Itamar Friedman, Matan
  Protter, and Lihi Zelnik-Manor.
\newblock Asymmetric loss for multi-label classification.
\newblock In {\em ICCV}, pages 82--91, 2021.

\bibitem{shao2019objects365}
Shuai Shao, Zeming Li, Tianyuan Zhang, Chao Peng, Gang Yu, Xiangyu Zhang, Jing
  Li, and Jian Sun.
\newblock Objects365: A large-scale, high-quality dataset for object detection.
\newblock In {\em ICCV}, pages 8430--8439, 2019.

\bibitem{tajbakhsh2016convolutional}
Nima Tajbakhsh, Jae~Y Shin, Suryakanth~R Gurudu, R~Todd Hurst, Christopher~B
  Kendall, Michael~B Gotway, and Jianming Liang.
\newblock Convolutional neural networks for medical image analysis: Full
  training or fine tuning?
\newblock {\em IEEE transactions on medical imaging}, 35(5):1299--1312, 2016.

\bibitem{tan2019learning}
Zhiyu Tan, Xuecheng Nie, Qi Qian, Nan Li, and Hao Li.
\newblock Learning to rank proposals for object detection.
\newblock In {\em ICCV}, pages 8273--8281, 2019.

\bibitem{tian2019fcos}
Zhi Tian, Chunhua Shen, Hao Chen, and Tong He.
\newblock Fcos: Fully convolutional one-stage object detection.
\newblock In {\em ICCV}, pages 9627--9636, 2019.

\bibitem{wang2022yolov7}
Chien-Yao Wang, Alexey Bochkovskiy, and Hong-Yuan~Mark Liao.
\newblock Yolov7: Trainable bag-of-freebies sets new state-of-the-art for
  real-time object detectors.
\newblock {\em arXiv preprint arXiv:2207.02696}, 2022.

\bibitem{xu2022pp}
Shangliang Xu, Xinxin Wang, Wenyu Lv, Qinyao Chang, Cheng Cui, Kaipeng Deng,
  Guanzhong Wang, Qingqing Dang, Shengyu Wei, Yuning Du, et~al.
\newblock Pp-yoloe: An evolved version of yolo.
\newblock {\em arXiv preprint arXiv:2203.16250}, 2022.

\bibitem{yang2016wider}
Shuo Yang, Ping Luo, Chen-Change Loy, and Xiaoou Tang.
\newblock Wider face: A face detection benchmark.
\newblock In {\em CVPR}, pages 5525--5533, 2016.

\bibitem{zhang2022dino}
Hao Zhang, Feng Li, Shilong Liu, Lei Zhang, Hang Su, Jun Zhu, Lionel~M Ni, and
  Heung-Yeung Shum.
\newblock Dino: Detr with improved denoising anchor boxes for end-to-end object
  detection.
\newblock {\em arXiv preprint arXiv:2203.03605}, 2022.

\bibitem{zhao2022omdet}
Tiancheng Zhao, Peng Liu, Xiaopeng Lu, and Kyusong Lee.
\newblock Omdet: Language-aware object detection with large-scale
  vision-language multi-dataset pre-training.
\newblock {\em arXiv preprint arXiv:2209.05946}, 2022.

\bibitem{zhao2020object}
Xiangyun Zhao, Samuel Schulter, Gaurav Sharma, Yi-Hsuan Tsai, Manmohan
  Chandraker, and Ying Wu.
\newblock Object detection with a unified label space from multiple datasets.
\newblock In {\em ECCV}, pages 178--193, 2020.

\bibitem{zhou2022simple}
Xingyi Zhou, Vladlen Koltun, and Philipp Kr{\"a}henb{\"u}hl.
\newblock Simple multi-dataset detection.
\newblock In {\em CVPR}, pages 7571--7580, 2022.

\bibitem{zhudeformable}
Xizhou Zhu, Weijie Su, Lewei Lu, Bin Li, Xiaogang Wang, and Jifeng Dai.
\newblock Deformable detr: Deformable transformers for end-to-end object
  detection.
\newblock In {\em ICLR}.

\end{thebibliography}
}

\clearpage
\appendix
\begin{center}
    \Large
    Supplimentary Materials
    \normalsize 
\end{center}

\setcounter{section}{0}
\setcounter{table}{0}
\setcounter{figure}{0}
\renewcommand\thesection{\Alph{section}}

\section{Comparisons on Multi-Label Classification}
As illustrated in Sec.4.3, our proposed CEM can perform the multi-label classification task. Therefore, in Table \ref{table:multi-label-classification} we make a comparison with the recent representative multi-label classification method, Query2Lable \cite{liu2021query2label}. METR is trained on COCO train2017 dataset while Query2Lable is trained on COCO 2014 benchmark, thus we reimplement it \cite{liu2021query2label} on COCO 2017. Compared with Query2label, our proposed CEM achieves favorable performance on the multi-label classification task. Specifically, CEM achieves 0.22 mAP gain with ResNet50 backbone \cite{he2016deep} and 448*448 resolutions. Following the traditional setting of object detection, the image with 800*1333 resolution is inputted and the performance of CEM is 89.36 mAP. It indicates that CEM can get relatively accurate multi-classification results, and thus can underpin the correctness of pre-defining categories for queries in advance. As for the image size of 448*448 and 800*1333, loading our pre-trained model can lead to improvements of 1.32 mAP and 0.75 mAP, respectively, which proves that the pre-trained model can boost the performance of multi-label classification. 


\section{Performance and Convergence Tendency}

\begin{figure}
\begin{center}
\includegraphics[width=0.95\linewidth]{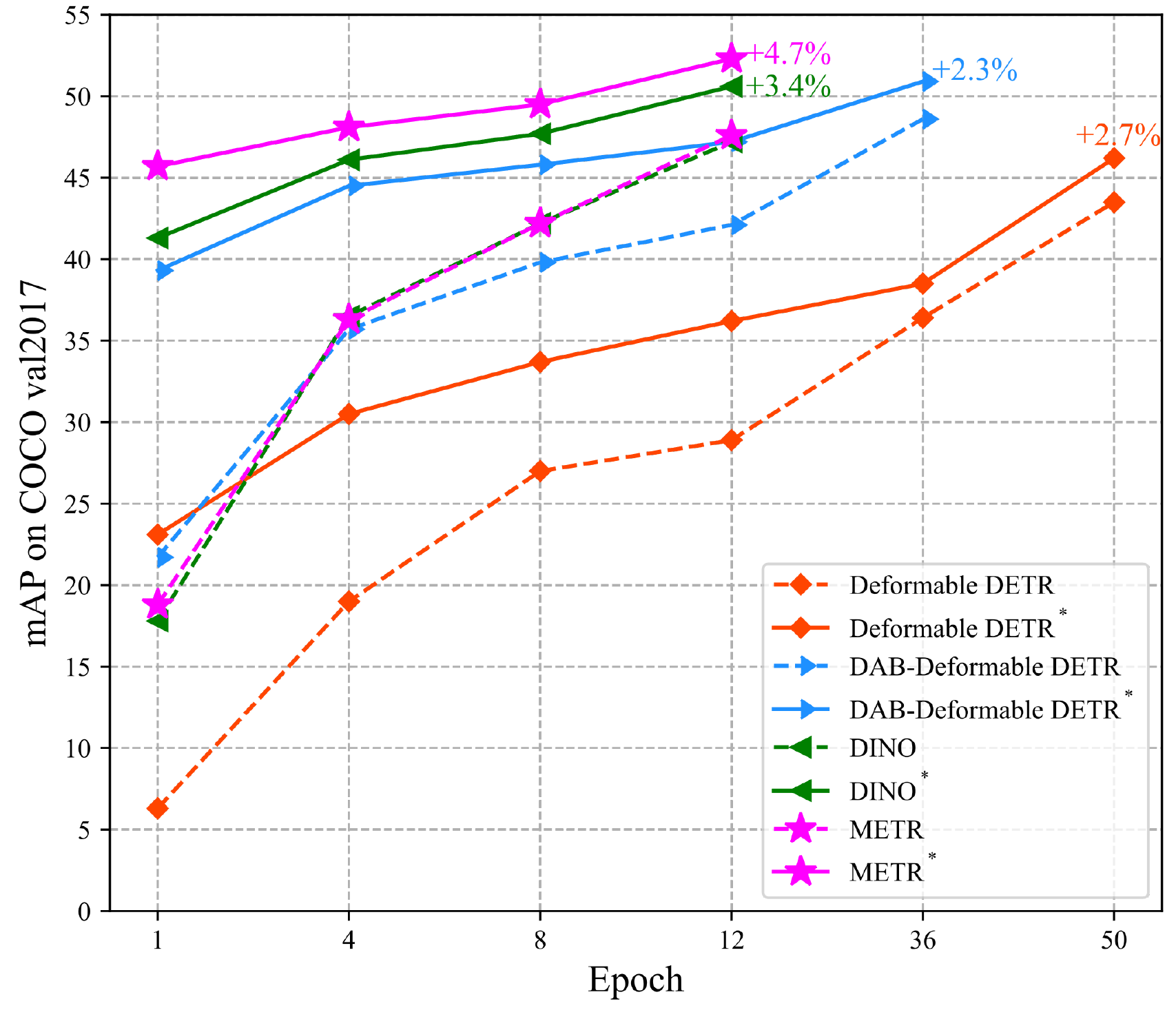}
\vspace{-14pt}
\end{center}
   \caption{The training convergence curves of different DETR-like detectors. * means loading our pre-trained model. Note that the curves of DINO \cite{zhang2022dino} and METR without * are closely overlapped.}
\label{fig:tendency}
\vspace{-6pt}
\end{figure}

We plot the convergence tendency of average precision on COCO val2017 upon various DETR-like detectors based on whether loading our METR pre-trained model. From Figure \ref{fig:tendency}, we can draw the conclusion that loading our pre-trained model can accelerate the training convergence obviously and improve the performance on average precision upon multiple DETR-like detectors. Taking METR as an example, the 1st epoch result with loading the pre-trained model almost is on par with the 12th epoch result of training with ImageNet-1K pre-trained ResNet50. In addition, our pre-trained model can dramatically increase the performance of various DETR-like models with different training schedules. These phenomenons indicate that our pre-trained model has highly flexible transferability and huge potential on boosting the performance of DETR-like detectors.
\vspace{-4pt}
\begin{table}[]
\centering
\caption{Experiments on COCO val2017 multi-label-classification. Note that Query2label is trained with 80 epochs while METR only is trained with 12 epochs. The ``IN1K'', ``O365'', and ``OImg'' refer to the ``ImageNet-1K'', ``Objects365'', and ``OpenImages'', respectively.}
\vspace{-10pt}
\begin{tabular}{p{1.4cm}p{1.2cm}<{\centering}p{1.8cm}<{\centering}p{1.2cm}<{\centering}p{0.8cm}<{\centering}}
\toprule
Method      & Backbone & Pretraining & Resolution & mAP   \\
\midrule
Query2Lable & R50      & IN1K                & 448*448    & 84.80 \\
Query2Lable & R101     & IN1K                & 448*448    & 85.97 \\
METR        & R50      & IN1K                 & 448*448    & 85.02 \\
METR        & R50      & OImg+O365           & 448*448    & \textbf{86.34} \\
METR        & R50      & IN1K                 & 800*1333   & 89.36 \\
METR        & R50      & OImg+O365          & 800*1333   & \textbf{90.11} \\
\bottomrule
\end{tabular}
\vspace{-5pt}
\label{table:multi-label-classification}
\end{table}

\begin{table*}[t]
\centering
\begin{tabular}{l}
\toprule
\textbf{Algorithm 1} Head Prediction of METR.    \\
\midrule
\textbf{Input:} Decoder layer output $Q_d$, Language embedding $E=[e^1, e^2,…,e^{TopK}] \in \mathbb{R}^{TopK \times d} $, \\ 
\ \ \ \ \ \ \ \ \ \ \ \ Reference point $(r_x, r_y, r_w, r_h)$ \\ 
 \textbf{Output:} Scores for TopK categories $S=\{s_i \in \mathbb{R}^{N}|, i=1,2,…,TopK\} \in \mathbb{R}^{(TopK*N)}$  \\
\ \ \ \ \ \ \ \ \ \ \ \ \ \ \ The corresponding bounding box $(x_c,y_c,w,h)$ \\
\textbf{Params:} The maximum number of category in one image $TopK$, \\ 
\ \ \ \ \ \ \ \ \ \ \ \ \ \ \ \ The number of object queries per class $N$, \\
\ \ \ \ \ \ \ \ \ \ \ \ \ \ \ \ Learnable bias $b$, \\
\ \ \ \ \ \ \ \ \ \ \ \ \ \ \ \ The multi-layer perception for class prediction $MLP_{cls} \in \mathbb{R}^{(TopK*N) \times 1}$, \\
\ \ \ \ \ \ \ \ \ \ \ \ \ \ \ \ The multi-layer perception for bounding box prediction $MLP_{box} \in \mathbb{R}^{(TopK*N) \times 4} $, \\
\ \ \ \ \ \ \ \ \ \ \ \ \ \ \ \ The language embedding dimension size $d$, \\
\ \ \ \ \ \ \ \ \ \ \ \ \ \ \ \  ${Sig}$ represents sigmoid function, ${Sig}^{-1}$ represents inverse sigmoid function. \\
 1: Calculate the class logits $P_{cls}$ and the bounding box logits $P_{box}$. \\
 2: $P_{cls} = MLP_{cls}(Q_d), P_{box} = MLP_{box}(Q_d)$. \\
 3: Calculate the Scores for TopK categories $S$. \\
 4: $S = P_{cls} * E / \sqrt{d} + b$. \\
 5: Calculate the bounding box. \\
 6: $(x_c,y_c,w,h) = {Sig}(P_{box} + {Sig}^{-1}(r_x, r_y, r_w, r_h))$. \\
 7: \textbf{return} Scores for TopK categories $S$, bounding box$(x_c,y_c,w,h)$. \\
\bottomrule
\label{algorithm}
\end{tabular}
\end{table*}
\vspace{0pt}
\begin{table}[!t]
\centering
\caption{Abalations on different learning rate strategies in the process of finetuning. ``S1'', ``S2'', and ``S3'' denote ``Strategy 1', ``Strategy 2'', and ``Strategy 3'', respectively. ``Back.'', ``Enc.'', and ``Dec.'' are the abbreviation of ``Backbone'', ``Encoder'', and ``Decoder'', respectively.}
\begin{tabular}{lp{0.8cm}<{\centering}p{0.8cm}<{\centering}p{0.8cm}<{\centering}p{0.8cm}<{\centering}p{0.6cm}<{\centering}p{0.6cm}<{\centering}}
\toprule
method & LR & Back. & Enc. & Dec.  & mAP  & AP50 \\
\midrule
& Default     & 1e-5     & 1e-4    & 1e-4      & 51.7 & 69.7 \\
METR &  S1  & 1e-5     & \textbf{2e-5}    & \textbf{2e-5}      & 52.1 & 69.9 \\
& S2  & 1e-5     & 2e-5    & \textbf{4e-5}      & 52.3 & 70.2 \\
& S3  & 1e-5     & \textbf{4e-5}    & 2e-5      & \textbf{52.4} & \textbf{70.3} \\
\midrule
DINO & Default     & 1e-5     & 1e-4    & 1e-4   & 50.7 & 68.0 \\
 & S3      & 1e-5     & 4e-5    & 2e-5   & 50.4 & 67.7 \\
\bottomrule
\end{tabular}
\label{table:Learning Rate Strategy}
\end{table}

\section{The Head Prediction Algorithm}
In Algorithm 1, we provide a detailed algorithm description about the part of head prediction of METR for a better understanding of the binary classification process, which is our core idea for solving the problem of label inconsistency among multiple object detection datasets. 

\section{Other Implementation Details}
\subsection{Detailed Model Components}
Following the common-used settings of DETR-like detectors, we exploit the 4-scale features for transformer encoder layers. 4-scale features come from stages 2, 3, and 4 of the backbone, and an extra feature is down-sampled from the output of stage 4. For query initialization, we follow DINO which initializes anchor boxes using the position information associated with the selected top-K features from the last encoder layer. The difference is that we initialize the content queries by language embeddings while DINO randomly initializes them. This is the main reason that METR possesses the ability of language-aware detection. Besides, we also adopt the contrastive denoising training strategy proposed by DINO for stabilizing training and accelerating convergence.
\vspace{-6pt}
\subsection{Learning Rate Setting for finetuning}
We explore the effect of the different learning rate (LR) settings during the finetune stage. The DETR-like detectors consist of three main parts, including the backbone, encoder, and decoder. The default setting of LR on the three parts is 1e-5, 1e-4, and 1e-4, respectively \cite{carion2020end, zhang2022dino, zhudeformable}. Pre-trained models usually have the ability to represent rich contextual semantics for images, thus it’s better to finetune the model to fit the customized dataset distribution with lower LR. We conduct extensive experiments on LR strategy, and the results are shown in Table \ref{table:Learning Rate Strategy}. Note that the LR of query embeddings and head predictions is 1e-4 in all experiments. We conclude that the LR of encoders and decoders should be decreased and the part of encoders’ LR should be slightly larger than decoders.
First, with our pre-trained model and default LR strategy, the mean average precision (mAP) of METR on COCO 2017val is 51.7. When we set the LR of both the encoder and decoder from 1e-4 to 2e-5 (Strategy 1), the mAP increases by 0.4\%. Compared with Strategy 1, separately enlarging the LR of encoders (Strategy 3) and decoders (Strategy 2) can improve the performance of 0.3\% and 0.2\%, respectively. It indicates that, in the process of finetuning, it’s more suitable that the LR of encoders is slightly larger than encoders.
We also apply the optimal Strategy 3 into DINO when finetuning DINO pre-trained models on Objects365, while it sightly drops 0.3\% mAP compared with its own baseline. We suspect that the reason why this phenomenon happened is the difference in the way of implementing object detection tasks. DINO predicts the classification and bounding boxes at the same time while METR predicts the categories involved in an image and then outputs the relative bounding boxes.

\section{Visualizations}
\begin{figure*}
\begin{center}
\includegraphics[width=0.95\linewidth]{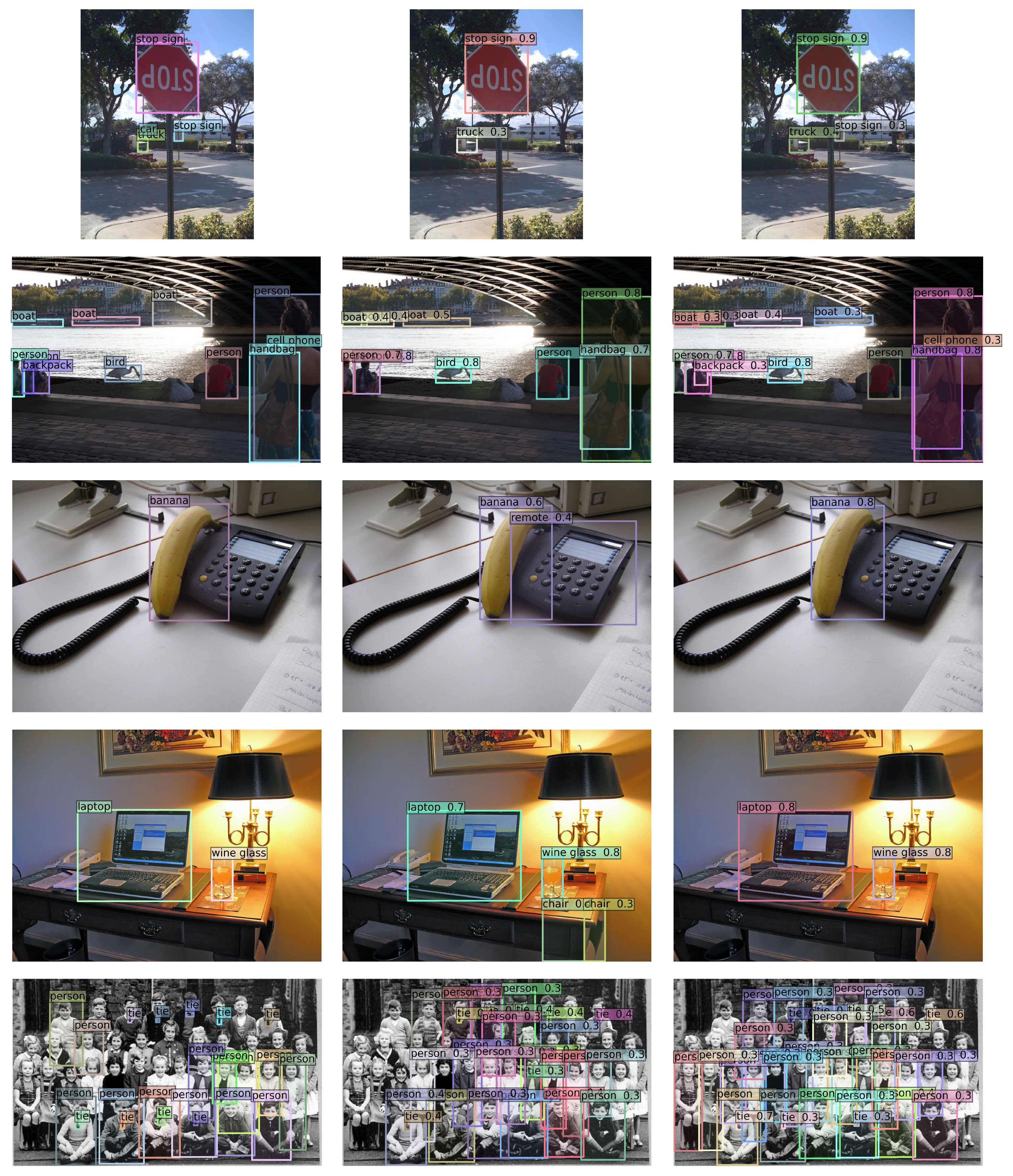}
\vspace{-6pt}
\end{center}
   \caption{Visualizations. From left to right, the figure shows the ground truth, predictions with DINO, and predictions with METR.}
\label{fig:Visualizations}
\end{figure*}

As shown in Table 2 of our paper, DINO loads its Object365 pre-trained model achieving 50.7 mAP and our proposed METR loads the joint pre-trained model (Objects365 and OpenImages) achieving 52.4 mAP, \ie, making 1.7 mAP gain. Here we make some visualizations in Figure \ref{fig:Visualizations}. The first column represents the ground truth, and the second and last columns show the predictions of DINO and METR, respectively. We find that the gain mainly comes from the following three points: 1) fewer missed detections: \eg, the ``stop sign'' and the ``boat''; 2) fewer incorrect detections: \eg, the ``remote'' and the ``chair''; 3) higher scores of detections: \eg, the ``banana''. Besides, we also find that for some objects where there is no label in the ground truth, METR can detect more of them than DINO (the last row in Figure \ref{fig:Visualizations}).

\end{document}